%% file: root.tex
\documentclass[letterpaper, 10 pt, conference]{ieeeconf}  

\IEEEoverridecommandlockouts                              

\usepackage{graphicx}
\usepackage{graphics} 
\usepackage{epsfig} 

\usepackage{amsmath} 
\usepackage{amssymb}  
\usepackage{algorithmic}
\usepackage{algorithm}
\usepackage{censor}
\usepackage{tabularx}
\usepackage{multirow}
\usepackage{color}
\usepackage{cite} 
\usepackage{url}
\usepackage{caption}
\usepackage{subcaption}
\usepackage{breqn}
\newcommand{\junk}[1]{}
\usepackage{algorithm}

\usepackage{dblfloatfix} 
\usepackage[export]{adjustbox}
\usepackage{tabulary,booktabs}
\usepackage{labelschanged}

\title{\LARGE \bf
HiL-ResRL: A Model-Agnostic Finetuning Adapter via Human-in-the-loop Residual Reinforcement Learning}
\author{
  Jingyi Liu$^{1}$, Zhaohong Mai$^{2}$, ShunSen He$^{1}$, Hang Ren$^{1}$, Chao Wang$^{1}$, Shunbo ZHOU$^{1}$, \\
  XiaoDong Wu$^{1}$, Heng Zhang$^{1,*}$
  \thanks{$^{1}$CloudRobo Lab, Huawei Cloud Computing Technologies Co., Ltd.}%
  \thanks{$^{2}$South China University of Technology. }%
  \thanks{$^{*}$Corresponding author}
}
\begin{document}

\maketitle
\thispagestyle{empty}
\pagestyle{empty}

\begin{abstract}
Recent advancements in generative imitation learning have significantly propelled the field of robotic manipulation. However, the majority of existing models rely heavily on Behavior Cloning (BC), a paradigm that suffers from compounding errors and distributional shift. Consequently, the efficacy of these models in practical industrial deployments remains limited. To address these challenges, we introduce a novel, plug-and-play fine-tuning pipeline designed to facilitate the robust deployment of Vision-Language-Action (VLA) models in real-world environments. In contrast to contemporary reinforcement learning (RL) fine-tuning strategies, which are often constrained by specific model architectures, our proposed framework is model-agnostic and adaptable to a diverse range of VLA models. We conceptualize VLA-generated actions as a unified interface, upon which we train a residual policy. This policy is designed to rectify suboptimal actions and address the distributional shift inherent in imitation learning. Additionally, we incorporate human-in-the-loop guidance to ensure safe exploration and maximize training efficiency. We conduct experiments directly in real-world robotic settings. The results demonstrate that within only 1.5 hour of real-world online RL training, the average success rate exceeds 95\% on real robots. Our work presents a practical solution for deploying behavior cloning models in industrial scenarios.
\end{abstract}

\input{intro.tex}
\input{rw.tex}
\input{algo.tex}

\input{results.tex}
\input{conclusions.tex}
\bibliographystyle{IEEEtran}
\bibliography{IROS_jingyi}
\end{document}

%% file: intro.tex
\section{INTRODUCTION}\label{Sec:intro}
Vision-Language-Action (VLA) models have proliferated rapidly and excel at generating plausible action sequences from multimodal inputs, offering a promising path toward generalist robotic control~\cite{openvla2024, black2026pi0visionlanguageactionflowmodel}. Despite their impressive capabilities, both visuomotor policies and VLA models rely predominantly on imitation learning (IL). This paradigm inherently suffers from the compounding error problem~\cite{asadi2019combatingcompoundingerrorproblemmultistep} and the distribution shift~\cite{nakamoto2025steeringgeneralistsimprovingrobotic}, which lead to degraded performance in real-world deployment~\cite{saxena2025matters}. Although increasing the number of real demonstrations or performing large-scale Sim-and-Real co-training can improve the success rate, both approaches exhibit performance saturation, making it difficult to meet the requirements for industrial deployment~\cite{maddukuri2025sim}.
 
To address these challenges, reinforcement learning (RL)~\cite{li2025reinforcement} has emerged as a common post-training solution to refine policies, correct systematic errors, and adapt to unseen environments. Existing work, such as RL4VLA~\cite{liu2026rlbringvlageneralization}, GRAPE~\cite{zhang2025grapegeneralizingrobotpolicy}, and SimpleVLA-RL~\cite{li2025simplevlarl}, have demonstrated improvements in the success rate and generalization of VLA models in simulation. Nevertheless, their performance often degrades in real-world deployment due to the sim-to-real gap. In contrast, real-world RL methods, such as $\pi^{*}_{0.6}$~\cite{intelligence2025pi06vlalearnsexperience}, achieve promising results in high-precision manipulation tasks by learning from autonomous experience and expert corrections. However, these approaches are computationally expensive due to extensive model parameter updates and are often tightly coupled with specific model architectures.

As flexible manufacturing advances, industrial factories’ expectations for production lines are shifting toward greater variety, shorter production cycles, and smaller batch sizes. This trend is particularly evident in high-end manufacturing, where rapid model deployment is increasingly required. In this context, balancing efficiency and performance in real-world deployment becomes critical. To address compounding errors and distribution shift, residual RL offers a promising model-agnostic direction~\cite{yuan2024policydecoratormodelagnosticonline}, but existing systems remain limited to simulation or suffer from low safety and efficiency during real-world training~\cite{he2025asapaligningsimulationrealworld}.

\begin{figure}[t]
    \centering
    \includegraphics[width=0.8\linewidth]{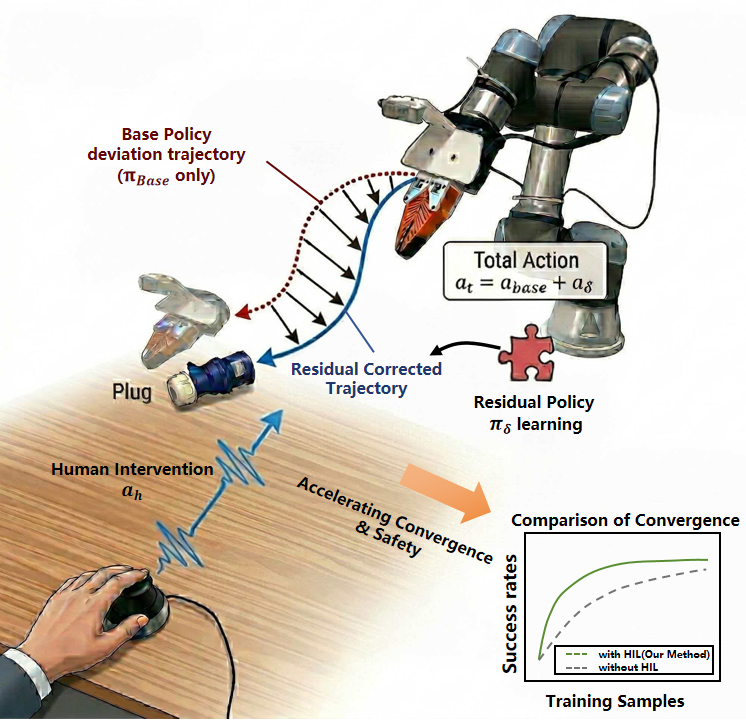}
    \caption{Overview of the Human-in-the-loop (HIL) online residual learning process.}
    \label{fig:HIL_overview}
\end{figure}

To enable efficient fine-tuning and strengthen training on hard-negative samples to mitigate deployment bottlenecks, human-in-the-loop (HIL) methods are introduced to provide expert-guided actions that accelerate training convergence. Thus, a human guided model-agnostic residual policy plugin is proposed for enhancing pre-trained VLA models. It treats various VLA as frozen, black-box action generators, employing a lightweight residual policy to compensate for systematic errors and distribution shifts. By integrating HIL into the RL training loop, the framework provides a robust safety layer for physical exploration while significantly enhancing sample efficiency, as shown in Fig.~\ref{fig:HIL_overview}. The main contributions of this work are summarized as follows:

\begin{itemize}
  \item \textbf{HIL-ResRL adapter:} We propose a novel HIL-ResRL adaptor, which serves as a flexible plugin compatible with both general visuomotor policies and VLA models. Notably, the residual network within the framework supports high-frequency reactive control, enabling real-time correction of the base policy’s actions and adaptive response to dynamic environmental changes.
  \item \textbf{Human-in-the-Loop Guided Online Learning:} We integrate human-in-the-loop guidance into the HIL-ResRL training process, constructing an online learning method based on human physical guidance and real-time feedback. This method facilitates safe exploration during real-world interactions while simultaneously streamlining human-robot skill transfer. 
  \item \textbf{Multi-modal Enhanced Residual Policy:} Beyond the vision-only HIL-ResRL adaptor, we extend the residual policy’s sensory inputs to include tactile/force feedback and verify that this enhancement improves performance on contact-rich tasks.
\end{itemize}

Experiments demonstrate that the proposed adapter enables behavior cloning policy to achieve superior real-world performance (over 95\%  success rate) in only 1 hour of online training, offering a practical and extensible final step for robust robotic deployment.

%% file: rw.tex
\section{Related Work}\label{Sec:rw}
\subsection{Residual Policy}
The residual policy was originally used to address the gradient vanishing problem in deep neural networks~\cite{he2016deep}, and this concept has now been widely applied in the field of robot control. Residual policies within the imitation learning (IL) paradigm have become a mainstream lightweight refinement approach for pre-trained robotic policies. Yuan et al.~\cite{yuan2024policydecoratormodelagnosticonline} presented a model-agnostic residual policy for online refinement of large IL models. TRANSIC~\cite{jiang2024transicsimtorealpolicytransfer} learned residual policies from human interventions to address sim-to-real gaps.
Compliant Residual DAgger~\cite{xu2025compliantresidualdaggerimproving} introduced a force-aware compliant residual policy that improves performance in contact-rich tasks. ASAP~\cite{he2025asapaligningsimulationrealworld} leverages a delta residual policy to compensate for unmodeled dynamic mismatches between simulation and the real world.

Despite their promising empirical results, these IL-based works lack sufficient robustness when confronted with environmental variations that fall outside the distribution of the demonstration data, leading to degraded performance in real-world scenarios.
To mitigate this limitation, residual policies have been further integrated with reinforcement learning (RL) methodologies.
ResFiT~\cite{ankile2025residualoffpolicyrlfinetuning} unified behavior cloning and off-policy RL within a residual framework for high-DoF robots.
PLD~\cite{xiao2025selfimprovingvisionlanguageactionmodelsdata} trained residual actors via RL to identify the failure regions of pre-trained VLA generalists. ResiP~\cite{ankile2025imitation} learned a closed-loop residual policy via RL to correct open-loop BC action chunks in real time. Although existing residual policy works have achieved remarkable progress in policy refinement, several critical limitations remain: the need for large amounts of real-robot data and the safety hazards introduced by random exploration, which destabilizes the base policy and caps task success rates. Building on residual
reinforcement learning, our work incorporates human-in-the-loop, which can intervene early in scenarios with high
failure rates, improve sample efficiency, effectively correct
cumulative errors, and ensure safety.

\subsection{Real-world RL}
Real-world reinforcement learning (RL) has emerged as a pivotal paradigm for endowing robots with adaptive manipulation capabilities.
One prominent line of research applies real-world RL to small policy models. Luo et al.~\cite{luo2024serl} propose SERL, a sample-efficient robotic RL software suite that enables the model to achieve near-perfect success rates on assembly tasks. Extending this framework, HIL-SERL~\cite{luo2025precise} integrates human demonstrations to further improve sample efficiency. Another direction explores real-world RL built upon VLA models, whose strong prior capabilities significantly accelerate early-stage exploration. Chen et al.~\cite{chen2025conrft} proposes the ConRFT framework, which alternates RL and SFT in real-world settings.  Zhai et al.~\cite{zhai2025visionlanguageactioncriticmodelroboticrealworld} develops the VLAC general process reward model to output dense progress signals for real-world RL, which eliminate task-specific reward engineering. 
Lei et al. \cite{lei2025rl100performantroboticmanipulation} propose RL-100, a three-stage pipeline for diffusion-based visuomotor policies, achieving exceptional success rates on diverse real-world tasks.
GR-RL~\cite{li2025grrlgoingdexterousprecise} combines offline data filtering, distributional critics, and online latent-space RL for high-precision dexterous manipulation tasks. $\pi^{*}_{0.6}$~\cite{intelligence2025pi06vlalearnsexperience} fine-tunes VLAs through the RECAP framework~\cite{intelligence2025pi06vlalearnsexperience} by using advantage-conditioned policies, leveraging both autonomous experience and expert corrections. 

Despite notable progress in real-world RL for robotic manipulation, several pervasive limitations remain unresolved, including safety-aware exploration, rapid training and fitting, and the fact that most existing frameworks are tailored to specific model frameworks or task types, thereby lacking model-agnostic and task-generalizable designs. Thus, the proposed HIL-ResRL adapter provides a plug-in module that can be seamlessly integrated with any VLA model, achieving both high efficiency and competitive performance.

%% file: algo.tex
\section{METHODS}\label{Sec:method}
\begin{figure*}[t]
    \centering
    \includegraphics[width=0.9\linewidth]{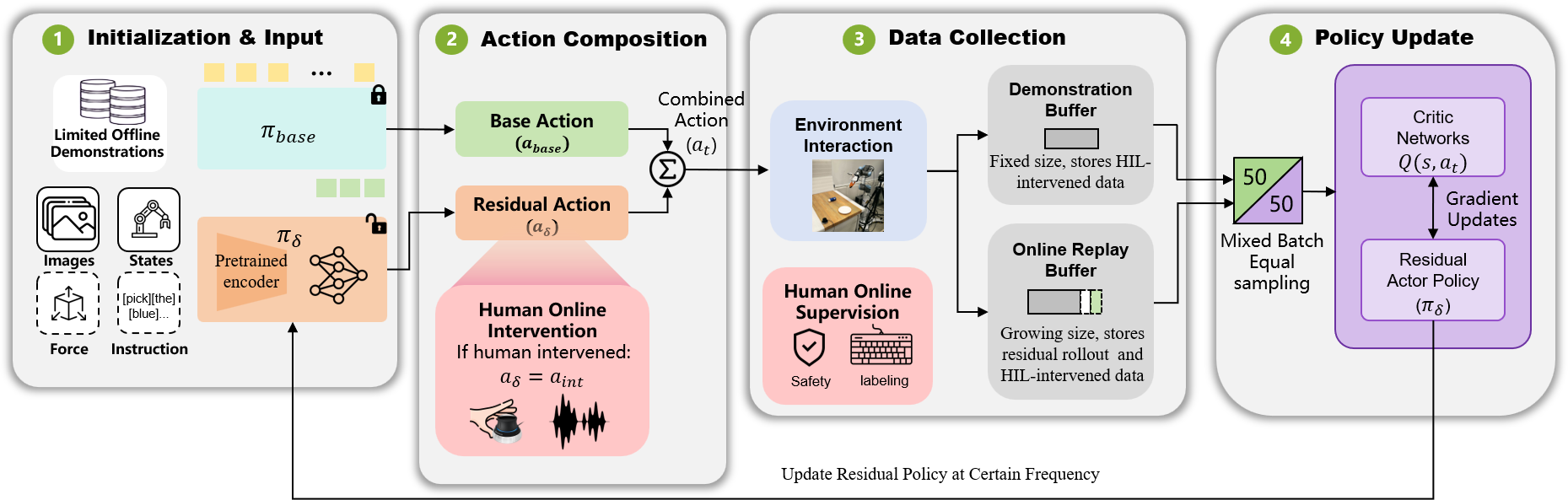}
    \caption{The Overview of HIL-ResRL.}
    \label{fig:overview}
\end{figure*}

In this section, we introduce a HIL-ResRL adapter that integrates residual reinforcement learning with human-in-the-loop interventions to enable efficient fine-tuning of the generative action policy. The overview of our method is illustrated in 
Fig.~\ref{fig:overview}

\subsection{Problem Formulation}

We formulate the robotic manipulation task as a Markov Decision Process, defined by the tuple $\mathcal{M} = (\mathcal{S}, \mathcal{A}, \mathcal{P}, \mathcal{R}, \gamma)$. At each timestep $t$, the state space ${s_t} = \left\{ {I_t^b,I_t^w,{p_t},{g_t}} \right\} \in \mathcal{S}$ consists of multi-view RGB images ${I_t^b}$ and ${I_t^w}$, proprioceptive state (6D TCP pose) ${p_t}$ and gripper width ${g_t}$. It chooses action ${a_t} \in \mathcal{A}$ by sampling from its policy $\pi \left( {{a_t}|{s_t}} \right)$. $\mathcal{P}$ is the transition dynamics, $\mathcal{R}$ is a function defining the rewards and ${\gamma}$ is the discount factor.

First, we train a behavior cloning policy, denoted as $\pi_{\psi}$ using a limited dataset of expert demonstrations ${\mathcal{D}_{sft}}$. This policy is designed to predict a sequence of $k$ future actions chunk $\mathbf{a}_{t:t+k-1}$, conditioned on the current observation $s_t$ and an optional language instruction $l$. The training objective is to maximize the log-likelihood of the demonstrations with $\tau = (s_t, l, \mathbf{a}_{t:t+k-1})$:

\begin{equation}
{\mathcal{L}_{BC}}(\psi) = - \mathbb{E}_{\tau \sim {\mathcal{D}_{sft}}} \bigl[ \log \pi_{\psi}(a_{t:t+k-1} \mid s_t, l) \bigr].
\end{equation} 
Then we employ the residual policy learning paradigm for reinforcement learning by decomposing the action into a base component $a_t^{base} \sim {\pi _{base}}({s_t}) = {\pi _\psi }({s_t})$, derived from imitation learning, and a residual component $a_t^{res}$:

\begin{equation}
    a_t = a_t^{base} + a_t^{res}
\end{equation}
where $a_t^{res} \sim \pi_{\theta}(s_t, a_t^{base})$ is the corrective action generated by the residual policy parameterized by $\theta$. For a trajectory horizon of length $H$, the training objective is to learn a policy $\pi_{\theta}$ that maximizes the expected cumulative reward as follow:

\begin{equation}
J(\theta) = \mathbb{E}[\sum_{t=0}^{H} \gamma^t r(s_t, a_t)]
\end{equation}where $r$ is a sparse binary reward indicating task success and ${\gamma ^t} \in [0,1]$ is the discount factor.


\subsection{Human-in-the-Loop Interventions}

To synergistically integrate human attributes, specifically intelligence and dexterity, with the trajectories generated by the Imitation Learning (IL) policy, we employ a human-in-the-loop mechanism. This approach aims to ensure safety-aware exploration and enhance overall efficiency in complex tasks, proceeding through three distinct phases:

\subsubsection{Phase-Switching Intervention}

\begin{figure}[h]
    \centering
    \includegraphics[width=0.9\linewidth]{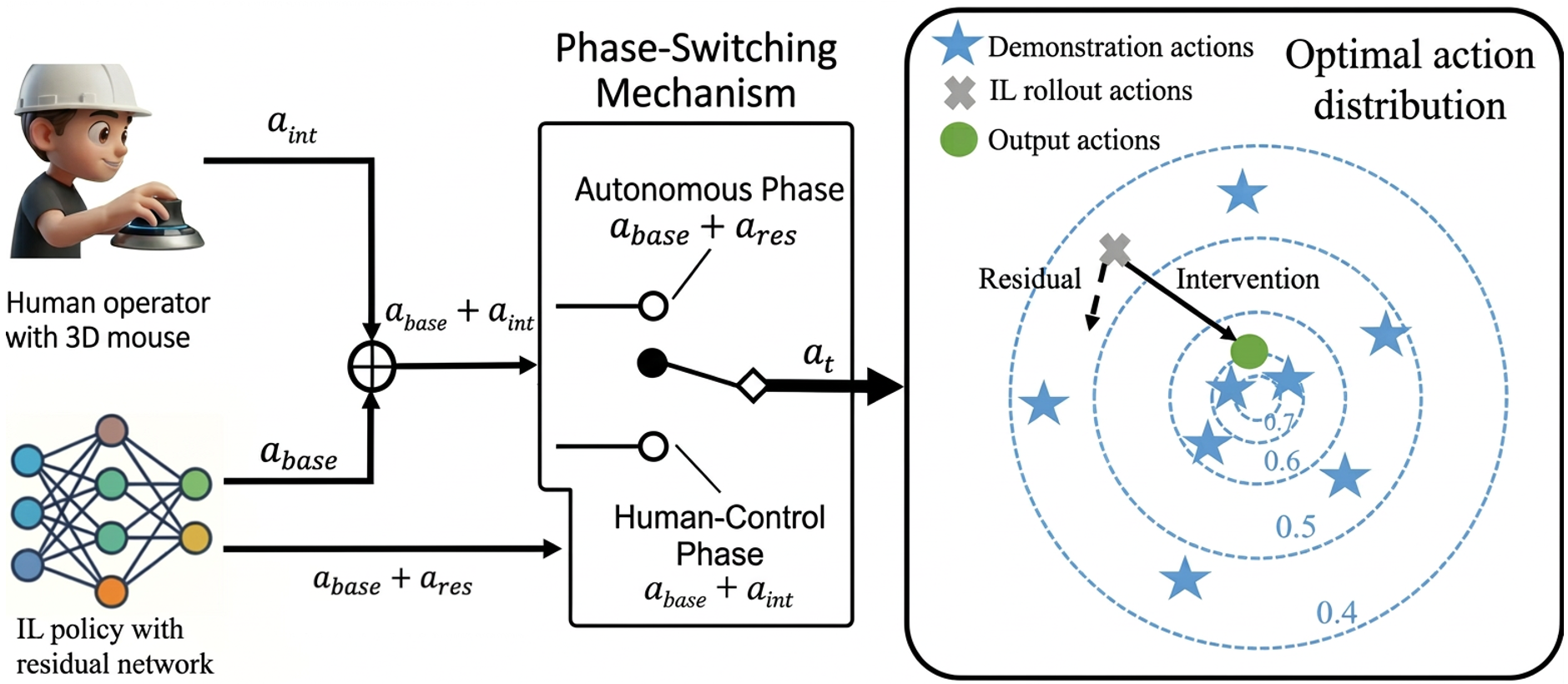}
    \caption{The scheme of human-in-the-loop intervention.}
    \label{fig:HIL_intv_overview}
\end{figure}

In the course of execution, the robot is susceptible to traversing regions where demonstration data is sparse, or it may encounter Out-of-Distribution (OOD) conditions. As depicted in Fig. 3, these anomalies typically result in oscillatory behavior or trajectory divergence. To mitigate this, a human operator is required to supervise the process online, providing a intervention signal ${x_t^{int}}$ during critical OOD state as follow:

\begin{equation}
a_t^{int} = \lambda  {x_t^{int}} 
\end{equation}
where ${\lambda}$ denote the scale factor between 0 and 1. The corrective input $a_t^{int}$ effectively steers the robot's state back towards a valid action probability distribution or an appropriate velocity gradient, which will be stored in the replay buffer $\mathcal{D}_{online}$ to guide the optimization. Rather than regenerating the entire trajectory (as shown in Fig. 3), this mechanism reduces the need for operating, thereby alleviating human's workload and stress.

\subsubsection{Safety Assurance and Success Judgment}

While automated vision classifiers \cite{luo2025precise} can predict task success, their reliability is limited for fine-grained manipulation tasks, particularly in industrial settings where occlusions can introduce significant visual ambiguity. As shown in Fig. 4, during the final stage of the plug-in-hole task, the base-camera view may fail to reveal that the plug is not fully inserted into the socket, whereas a human operator can make this judgment accurately. Moreover, to recognize physically hazardous states and subtle force-limit violations, human supervision serves two critical functions:

\begin{figure}[h]
    \centering
    \includegraphics[width=0.9\linewidth]{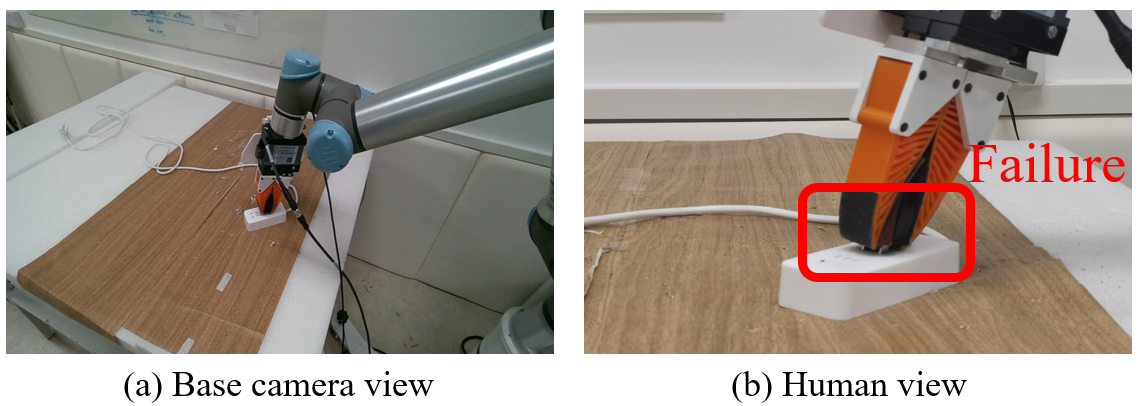}
    \caption{The plug-in-hole images from different views.}
    \label{fig:judgement_overview}
\end{figure}

\begin{itemize}
    \item \textbf{Success/Failure Labeling:} We utilize human's labels as the ground truth for task termination.
    \item \textbf{Emergency Reset:} If the robot enters a dangerous state or a deadlock that cannot be corrected via intervention, the operator triggers an immediate failure signal. This forces the environment to reset, ensuring hardware safety and preventing the policy from learning from hazardous states.
\end{itemize}

\subsubsection{Failure-aware Training}
By analyzing failure cases in terms of state distributions (e.g., out-of-distribution states or image occlusions), human-in-the-loop hard-negative mining can be used to augment training, thereby forcing the residual policy to undergo targeted learning in the most challenging regions of the state space.

\subsection{Residual Learning with Off-Policy RL}

To enable sample-efficient real-world training, we employ an off-policy RL framework. Specifically, we adopt the Soft Actor-Critic (SAC) algorithm, tailored to the residual learning setting, and incorporate layer normalization in the critic networks as well as ensembled double-Q learning. The complete training procedure for our Human-in-the-Loop Residual RL is summarized in Algorithm \ref{alg:hil_resrl}.


\subsubsection{Data Collection and Initialization}
Prior to residual training, we construct a demonstration dataset $\mathcal{D}_{demo}$ to bootstrap the learning process. We deploy the frozen base policy $\pi_{base}$ to interact with the environment. A human operator monitors the execution and intervenes only when necessary.
\begin{itemize}
    \item \textbf{Autonomous Phase:} When $\pi_{base}$ performs adequately, the human provides no input. The implied residual is zero ($a_{res} = 0$), and the effective action is $a_t = \pi_{base}(s_t)$.
    \item \textbf{Intervention Phase:} When the base policy deviates or fails, the human provides a corrective intervention $a_{int}$. The recorded residual action is effectively the human's correction relative to the base frame.
\end{itemize}
The collected transitions, including states, total actions (or effective residuals), and rewards, are stored in $\mathcal{D}_{demo}$. This dataset initializes the demonstration buffer for the subsequent RL phase.

\subsubsection{Human-in-the-Loop Training Process}
During the residual training phase, we maintain two distinct replay buffers: a fixed demonstration buffer $\mathcal{D}_{demo}$ and a growing online replay buffer $\mathcal{D}_{online}$. 

With the learned residual $a_{res} \sim \pi_{\theta}(\cdot|s_t, \pi_{base}(s_t))$ and human intervention $a_{int}$ via the teleoperation device. The executed action $a_t$ is determined by:

\begin{equation}
    a_t = 
    \begin{cases} 
    \pi_{base}(s_t) + \lambda a_{int} & \text{if ${a_{int}} \ne 0$} \\
    \pi_{base}(s_t) + \lambda a_{res} & \text{otherwise}
    \end{cases}
\end{equation}
where ${\lambda}$ denote the scale factor. \\
With the buffer storage strategy, transitions involving human interventions $((s_t, a_t, r_t, s_{t+1}))$ are stored in \textit{both} $\mathcal{D}_{demo}$ and $\mathcal{D}_{online}$ to explicitly guide the policy towards high-quality behaviors. Transitions generated purely by the residual policy are stored \textit{only} in $\mathcal{D}_{online}$. This mechanism ensures that $\mathcal{D}_{demo}$ remains a repository of high-quality corrective behaviors \cite{luo2025precise}. We introduce a scaling coefficient \(\lambda \in [0,1]\) to constrain the magnitude of the residual policy's output, preventing it from excessively overriding the base policy's actions. This restricts learning to local refinements, enabling a lightweight editor and facilitating more efficient and stable optimization.

\subsubsection{Parameter Optimization}
We update the policy parameters using a symmetric sampling strategy, drawing a batch of transitions $\mathcal{B}$ composed of $50\%$ samples from $\mathcal{D}_{demo}$ and $50\%$ from $\mathcal{D}_{online}$, referenced by \cite{luo2025precise}.

The critic minimizes the Bellman error over this mixed batch. We utilize an ensemble of $N$ critics $\{Q_{\phi_i}\}_{i=1}^N$ with Layer Normalization applied to the linear layers. The loss function for each critic is:
\begin{equation}
    \mathcal{L}_Q(\phi_i) = \mathbb{E}_{(s, a, r, s') \sim \mathcal{B}} \left[ \left( Q_{\phi_i}(s, a) - y \right)^2 \right]
\end{equation}
where the target $y$ is computed using the minimum of a subset of target critics (Double-Q Ensembled) and includes the entropy term $\alpha$:
\begin{equation}
\begin{aligned}
y = r + \gamma \biggl[ & \min_{j \in \mathcal{K}} Q_{\phi'_j}(s', \pi_{base}(s') + \tilde{a}_{res}') \\
& - \alpha \log \pi_{\theta}(\tilde{a}_{res}'|s', \pi_{base}(s')) \biggr]
\end{aligned}
\end{equation}

The residual actor $\pi_{\theta}$ is updated to maximize the expected Q-value of the hybrid action (base + residual) while maximizing entropy. The loss is calculated over the same mixed batch $\mathcal{B}$:

\begin{equation}
\begin{aligned}
\mathcal{L}_\pi(\theta)
&= \mathbb{E}_{s\sim\mathcal{B}} \left[
    \alpha \log \pi_\theta(\tilde a_{\mathrm{res}}\mid s,\pi_{\mathrm{base}}(s))
\right.\\
&\phantom{=} - \min_{j=1,2} Q_{\phi_j}\!\left(s,\pi_{\mathrm{base}}(s)+\tilde a_{\mathrm{res}}\right)
\left.\right]
\end{aligned}
\end{equation}

In our implementation, the residual policy is parameterized by a multilayer perceptron (MLP) with three hidden layers, each of width 256.

\begin{algorithm}[h]
\caption{Human-in-the-Loop Residual RL Training}
\label{alg:hil_resrl}
\begin{algorithmic}[247]
\STATE \textbf{Initialize:} Frozen base policy $\pi_{base}$, Residual policy $\pi_{\theta}$, Critics $Q_{\phi}$, Replay Buffers $\mathcal{D}_{demo}, \mathcal{D}_{online}$
\STATE \textbf{Collect Demos:} Run $\pi_{base}$ with human oversight; store human corrections and successes/failures in $\mathcal{D}_{demo}$
\FOR{each episode}
    \STATE Reset environment, observe $s_0$
    \FOR{each step $t$}
        \STATE Compute base action: $a_{base} \leftarrow \pi_{base}(s_t)$
        \STATE Sample residual: $a_{res} \sim \pi_{\theta}(\cdot | s_t, a_{base})$
        \IF{Human Intervention Detected}
            \STATE Get human action $a_{int}$ via teleoperation
            \STATE Execute $a_t \leftarrow a_{base} + a_{int}$
            \STATE Store $(s_t, a_t, r_t, s_{t+1})$ in $\mathcal{D}_{demo}$ \textbf{and} $\mathcal{D}_{online}$
        \ELSE
            \STATE Execute $a_t \leftarrow a_{base} + a_{res}$
            \STATE Store $(s_t, a_t, r_t, s_{t+1})$ in $\mathcal{D}_{online}$
        \ENDIF
        \STATE \textbf{Update Step:}
        \STATE Sample batch $b_{demo} \sim \mathcal{D}_{demo}$ and $b_{online} \sim \mathcal{D}_{online}$
        \STATE Combine $\mathcal{B} \leftarrow b_{demo} \cup b_{online}$ (50/50 mix)
        \STATE Update Critic $Q_{\phi}$ using Eq. (16) \& (17) with $\mathcal{B}$
        \STATE Update Actor $\pi_{\theta}$ using Eq. (18) with $\mathcal{B}$
        \STATE Update target networks
    \ENDFOR
\ENDFOR
\end{algorithmic}
\end{algorithm}

%% file: results.tex
\section{Experiments and Results}

\begin{figure*}[t]
    \centering
    \begin{subfigure}[b]{0.32\textwidth}
        \centering
        \includegraphics[width=\textwidth]{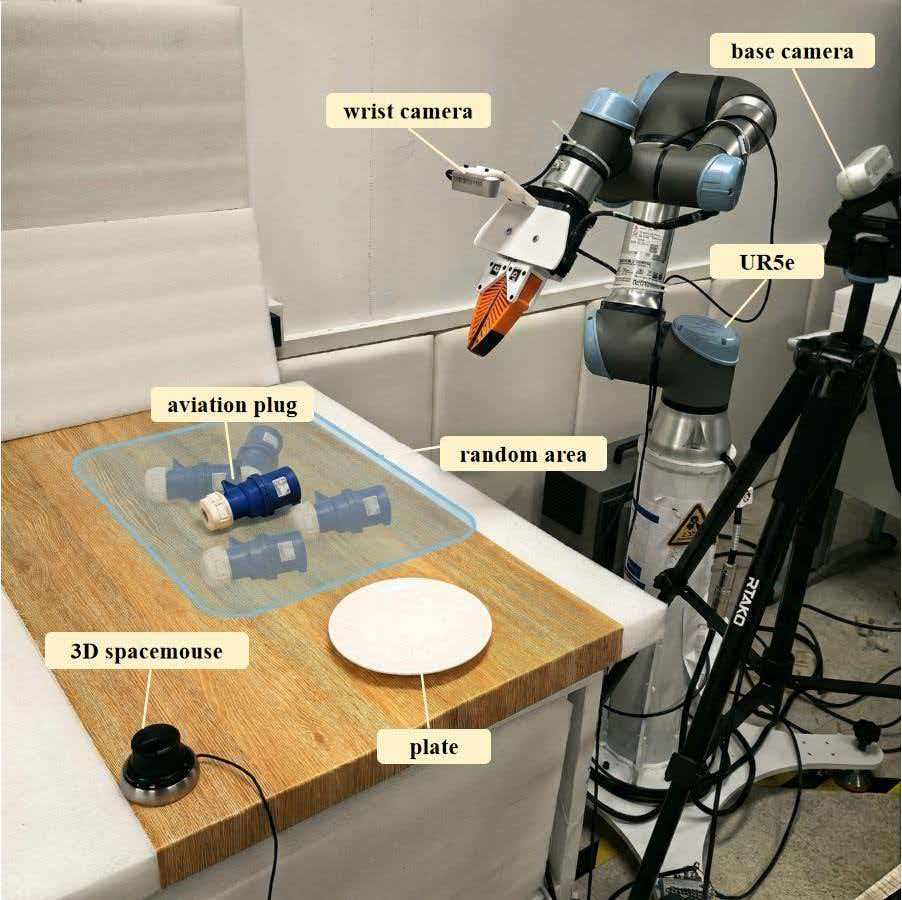}
        \caption{Pick and Place.}
        \label{fig:pick_place}
    \end{subfigure}
    \hfill
    \begin{subfigure}[b]{0.32\textwidth}
        \centering
        \includegraphics[width=\textwidth]{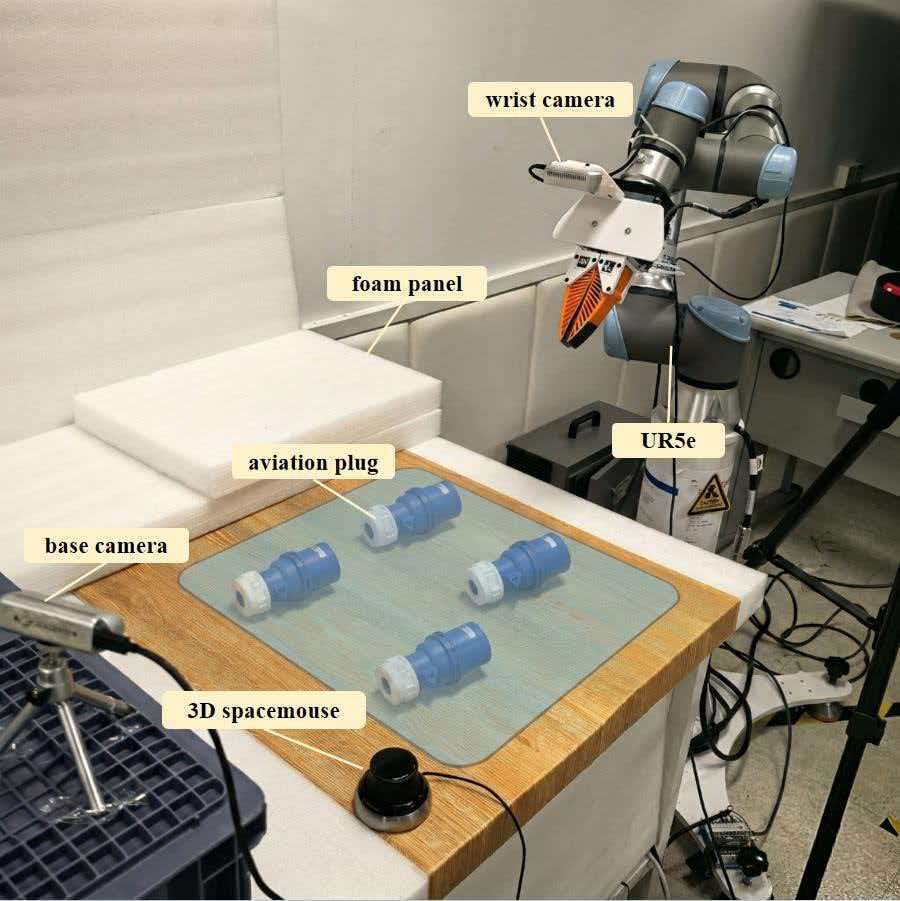}
        \caption{Place Upright.}
        \label{fig:place_upright}
    \end{subfigure}
    \hfill
    \begin{subfigure}[b]{0.32\textwidth}
        \centering
        \includegraphics[width=\textwidth]{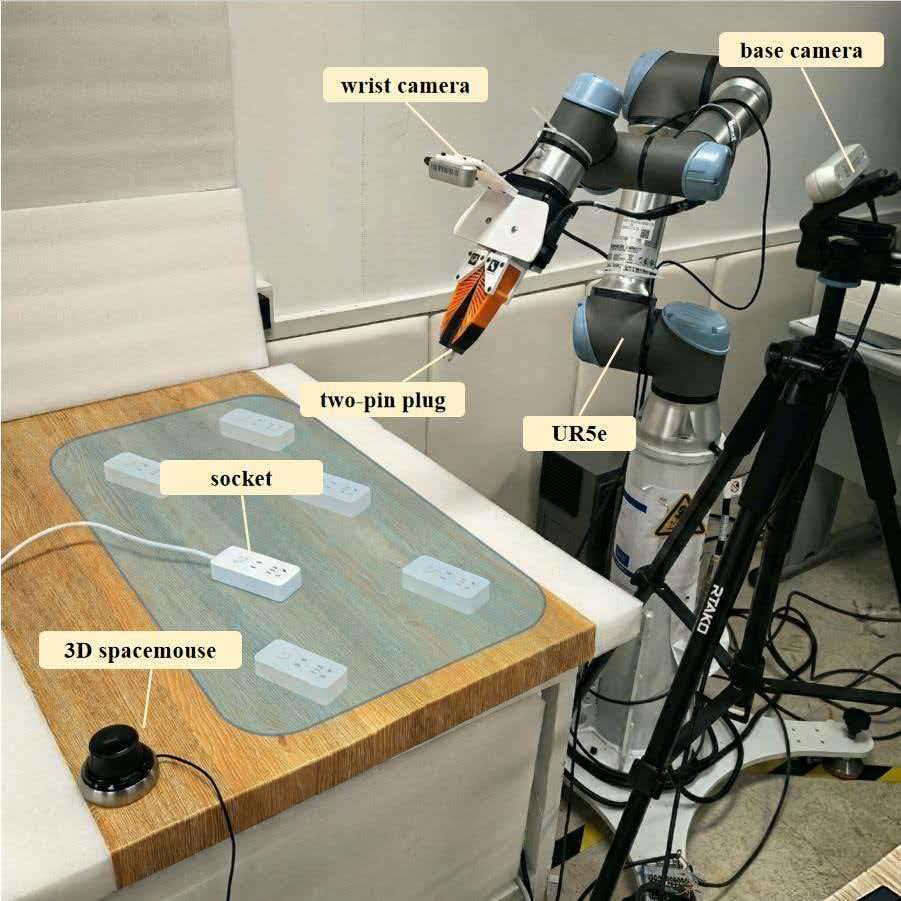}
        \caption{Plug in Hole.}
        \label{fig:plug_in_hole}
    \end{subfigure}
    \caption{Experimental setups for the three robotic manipulation tasks using the UR5e platform. (a) illustrates the pick and place setup, (b) shows the upright placement with foam panels, and (c) demonstrates the precision plugging task.}
    \label{fig:experimental_setups}
\end{figure*}

\label{sec:experiments}

To evaluate the generality, and efficiency of the proposed HIL-ResRL adapter, our experiments are designed to answer the following core research questions:

\begin{enumerate}
    \item \textbf{Q1 (Performance):} Can the proposed HIL-ResRL adapter consistently improve success rates across diverse tasks when paired with different base policies, and mitigate the real-world performance saturation observed with the base policies alone?
    \item \textbf{Q2 (Human-in-the-loop):} Is HIL intervention necessary for efficient residual learning, and does incorporating HIL improve sample efficiency on complex tasks?
    \item \textbf{Q3 (Contact-rich):} Can HIL-ResRL incorporate force/torque feedback to address contact-rich manipulation, and does it exhibit extensibility to multimodal inputs?
\end{enumerate}

To answer these questions, we conducted extensive evaluations mainly in the real-world deployments.

\subsection{Experimental Setup}

\subsubsection{Real-World Experiment Setup}
A collaborative robotic arm (UR5e), equipped with a UMI-like gripper and a 6-DoF force/torque sensor, is used to deploy the proposed system. The visual perception system comprises two Intel RealSense D435i RGB-D cameras: one mounted on the end effector to provide a first-person view and the other positioned laterally to provide a global view. Human interventions are provided via a 3D SpaceMouse.

To validate the proposed real-world RL framework, we design three representative tasks derived from industrial scenarios to systematically evaluate its performance, as illustrated in Fig.~\ref{fig:experimental_setups}. These tasks pose distinct challenges in dexterity and multimodal perception, as summarized below:

\begin{itemize}
    \item \textbf{Pick and Place:} In this task, the robot grasps an aviation plug from a random initial pose and places it at a designated target location. The task demands high precision in both grasping and placement across varying object poses.
    \item \textbf{Place Upright:} This task introduces an additional orientation constraint: after grasping the plug, the robot must reorient it to an upright, vertical posture before placing it stably on the target surface. This requirement necessitates more precise object-level manipulation and orientation control.
    \item \textbf{Multiple Plug-in-Hole:} Since the socket is partially occluded in both camera views during the plug-in-hole process, the robot must rely primarily on end-effector force/torque signals to achieve precise insertion. This setting evaluates the effectiveness of residual learning in exploiting force-domain information when vision alone is insufficient. Moreover, the multiple plug-in-hole task requires fine-grained control of the plug pose. Otherwise, it is difficult to align with the multiple apertures or the insertion may remain incomplete, as shown in Fig. 4.
\end{itemize}

\begin{figure*}[t]
    \centering
    \begin{subfigure}[b]{0.8\textwidth}
        \centering
        \includegraphics[width=\textwidth]{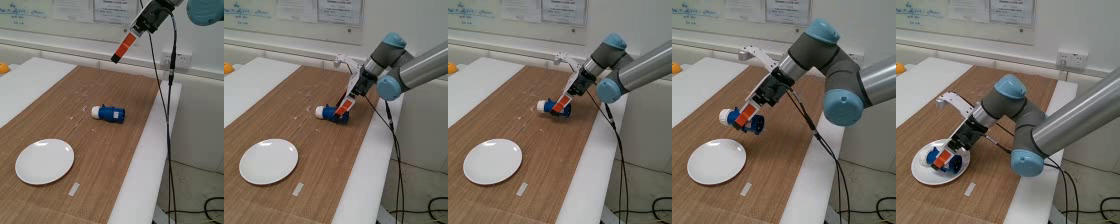}
        \caption{Pick and place process.}
        \label{fig:pick_place_process}
    \end{subfigure}
    
    \begin{subfigure}[b]{0.8\textwidth}
        \centering
        \includegraphics[width=\textwidth]{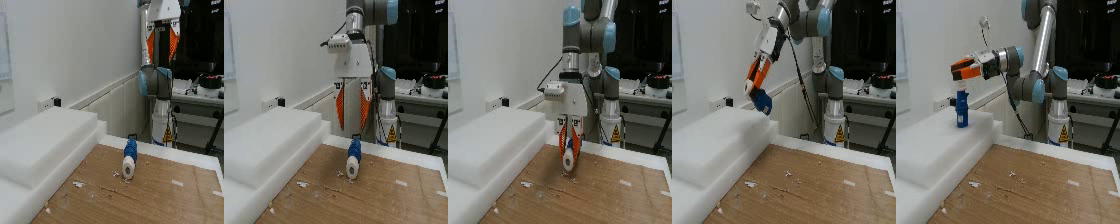}
        \caption{Upright placement process.}
        \label{fig:place_upright_process}
    \end{subfigure}
    
    \begin{subfigure}[b]{0.8\textwidth}
        \centering
        \includegraphics[width=\textwidth]{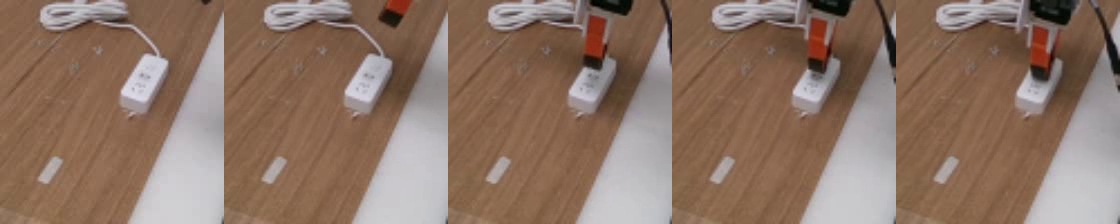}
        \caption{Precision plugging process.}
        \label{fig:plug_in_hole_process}
    \end{subfigure}
    \caption{Experimental processes for the three robotic manipulation tasks using the UR5e platform. (a) illustrates the pick and place process, (b) shows the upright placement process with foam panels, and (c) demonstrates the precision plugging process.}
    \label{fig:experimental_process}
\end{figure*}

To demonstrate the plug-and-play capability of proposed the HIL-ResRL adapter, we select representative and widely used methods from the visuomotor policy and VLA model as base policies, namely Diffusion Policy \cite{chi2023diffusionpolicy} and the flow-based  $\pi_{0.5}$ model \cite{intelligence2025pi05visionlanguageactionmodelopenworld}. We collect 50 expert demonstrations for the pick-and-place and place-upright tasks, and 80 demonstrations for the multiple peg-in-hole task, to train the base policies. This training ensures that each base policy attains a reasonable success rate, thereby providing a prior for subsequent residual exploration.

To enable low-latency control and efficient training, a distributed computing architecture is adopted for real-world RL. Specifically, inference for the base policy and the residual network is co-located on a workstation, while residual-network training is executed separately on another workstation.

\subsubsection{Baselines}
To address the three key questions posed above and to ablate the contribution of each component in HIL-ResRL, we design three comparative baselines:
\begin{itemize}
    \item \textbf{Behavior Cloning} The base policy (e.g., Diffusion Policy \cite{chi2023diffusionpolicy} or $\pi_{0.5}$ \cite{intelligence2025pi05visionlanguageactionmodelopenworld}) trained solely on offline demonstrations without any residual RL fine-tuning.
    \item \textbf{ResRL (w/o HIL):} We consider a residual RL baseline that operates without human intervention, inspired by reinforcement learning as explored in RLPD \cite{ball2023rlpd}. This baseline employs a sample-efficient off-policy RL algorithm that leverages prior data, using the same demonstration dataset as HIL-ResRL.
    \item \textbf{HIL-SERL \cite{luo2025precise}:} A state-of-the-art human-in-the-loop real-world RL framework that learns a policy from scratch via off-policy RL with human interventions, but does not exploit the structured prior provided by a frozen VLA base policy.
\end{itemize}

\subsection{Results and Analysis}

\begin{figure}[htbp]
    \centering
    \includegraphics[width=\linewidth]{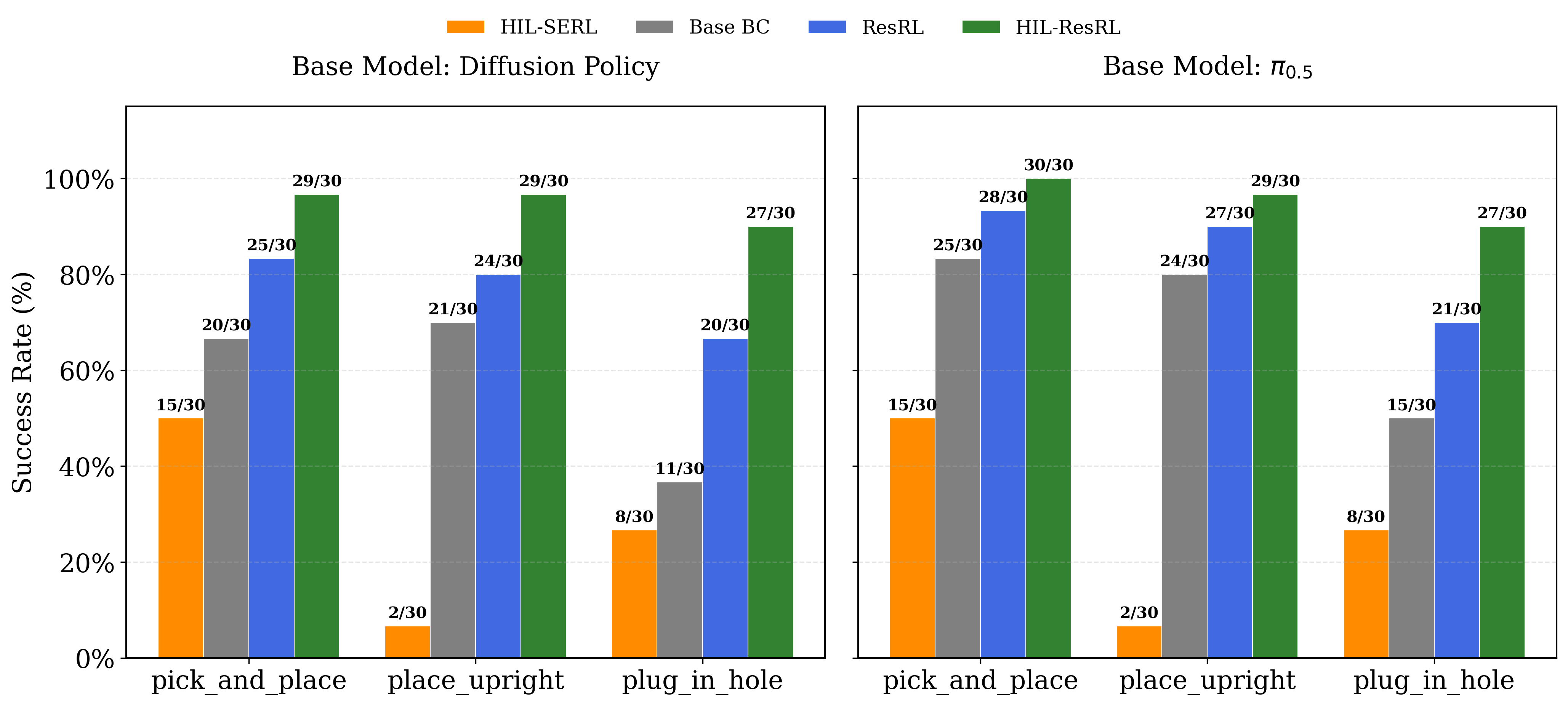}
    \caption{Real-world success rates on three tasks using Diffusion Policy (left) and $\pi_{0.5}$ (right) as base policies. For each task, bars show the performance of the base BC policy after SFT, the HIL-SERL baseline, and our HIL-ResRL. The online training time for each task is limited to 1.5 hours.}
    \label{fig:real_success}
\end{figure}

\subsubsection{Improving Task Success Rates (Q1)}
As shown in Fig. 7, the two base policies achieve success rates of 50\%–80\% across the three tasks. However, due to bottleneck states (e.g., tight insertions and precise grasps), the policies exhibit performance saturation. With HIL-ResRL, we increase the success rate to over 90\% within 40-90 minutes of training.

Furthermore, we observe that HIL-SERL exhibits significant limitations in tasks necessitating simultaneous, precise control of both end-effector position and orientation (e.g., place-upright and plug-in-hole). The HIL-SERL baseline attempts to learn the entire trajectory distribution, resulting in poor convergence and plateauing at suboptimal performance levels given the same training duration. In contrast, HIL-ResRL utilizes the base policy as an action prior to constrain exploration to local dynamics corrections, thereby obviating the need to explore the entire workspace.

To further investigate whether enhanced training with the HIL-ResRL adapter can address base policy instability or failure in hard-negative cases, we conducted generalization experiments regarding the target object pose. The comparison of success rates is presented in Fig. 8. These results further demonstrate that failure-aware training effectively resolves the distribution shift of the robot state relative to the demonstration data. Meanwhile, under identical experimental conditions, incorporating the HIL-ResRL adapter consistently yields higher task success rates than using the base policy alone. This indicates that HIL-ResRL preserves the strong capabilities of the base policy while providing a substantial performance improvement.

\begin{figure}[t]
    \centering
    \includegraphics[width=0.8\linewidth]{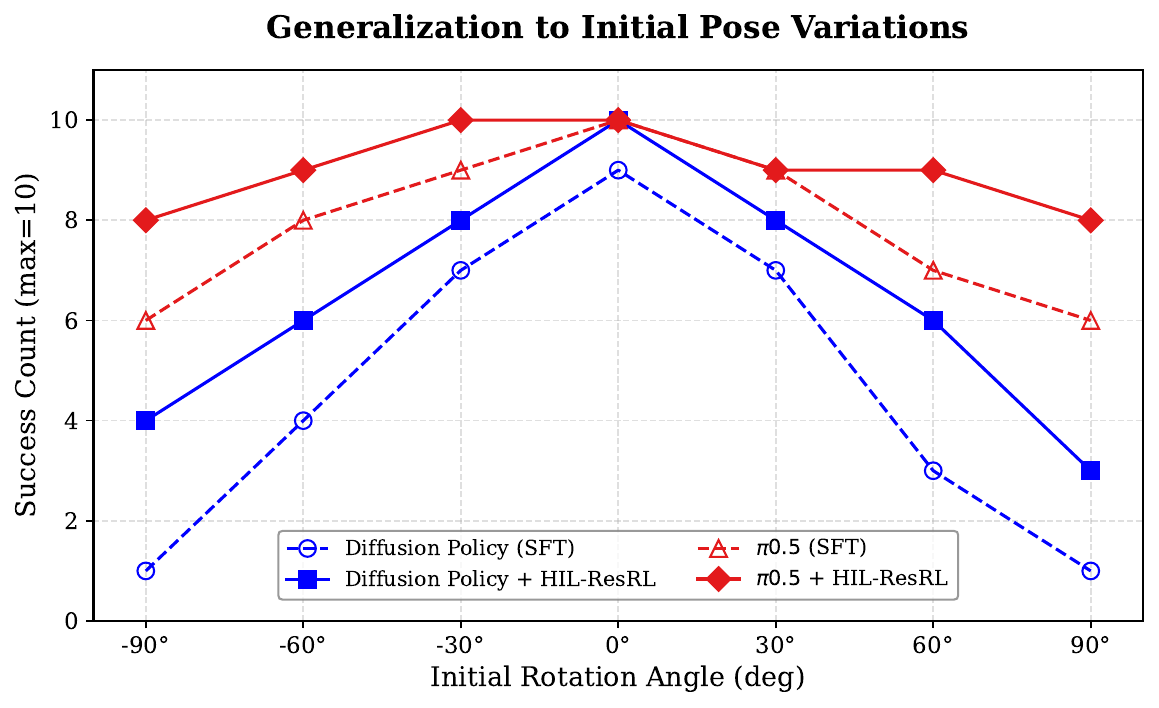}
    \caption{Generalization to initial plug orientation in the \texttt{place\_upright} task. Success rates are shown for Diffusion Policy and $\pi_{0.5}$ with or without HIL-ResRL adaptor, as a function of the initial rotation angle.}
    \label{fig:generalization}
\end{figure}

\subsubsection{Human-in-the-loop (Q2)}
To assess the necessity of human intervention, we conducted an ablation study comparing two training paradigms: HIL-ResRL and ResRL. With the HIL-ResRL adapter, beyond online residual corrections, we incorporated triggered reset signals that initiate a safe reset routine whenever the robot becomes irretrievably stuck. Conversely, the ResRL paradigm relies on fully autonomous exploration. As illustrated in Fig. 7, the experimental results demonstrate that HIL-ResRL achieves a significantly higher success rate than ResRL, attributable to the integration of human-in-the-loop interventions during training.

Notably, in both paradigms, automatic emergency stops may still be triggered by joint limits or excessive external forces. This challenge is particularly acute in the multiple plug-in-hole task, where the robot must not only maintain robust action inference across diverse pose states but also prevent high-force collisions when visual occlusion hinders proper alignment. For effective safety-aware training, minimizing emergency stops is imperative to preserve training continuity and mitigate hardware risks. As summarized in Fig. 9, a comparative analysis of the plug-in-hole experiments reveals that the autonomous baseline triggered 15 emergency stops within one hour of training, whereas HIL-ResRL incurred only 2. These results confirm that active human guidance significantly enhances safety by preventing the robot from entering hazardous regions of the state space.

\begin{table}[htbp]
  \centering
  \caption{Safety incidents during training}
  \label{tab:safety_incidents_training}
  \begin{tabular}{lc}
    \toprule
    Method & Emergency Stops \\
    \midrule
    ResRL (w/o human) & 15 \\
    HIL-ResRL (with human) & 2 \\
    \bottomrule
  \end{tabular}

  \vspace{0.5em}
\end{table}



\subsubsection{Effect of contact-rich task (Q3)}
\begin{figure}[htbp]
    \centering
    \includegraphics[width=0.9\linewidth]{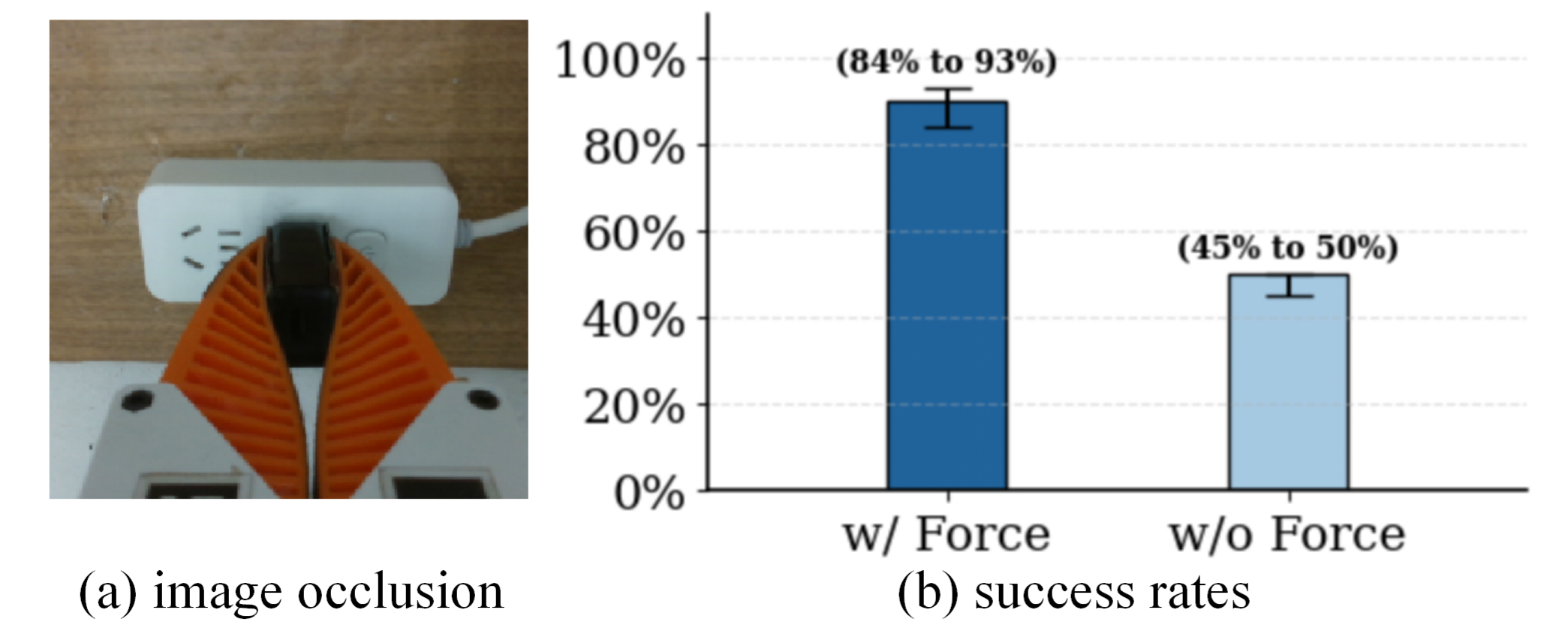}
    \caption{Visual occlusion analysis and performance comparison. The left panels (a) show the end-effector obstructing the socket during insertion. The right panel (c) quantifies the success rate with and without force feedback (93\% vs 50\%).}
    \label{fig:force_ablation}
\end{figure}

To evaluate the significance of haptic feedback, we compare two variants of HIL-ResRL: a force-aware model incorporating 6-axis end-effector F/T measurements and a baseline relying solely on visual and proprioceptive inputs. As illustrated in Fig.~\ref{fig:force_ablation}(b), the force-integrated variant achieves a significantly higher success rate (93\%) compared to the baseline (50\%).

We observe that while the base policy can guide the robot to the vicinity of the target, the absence of force feedback leads to "stalling" or excessive contact forces during the insertion phase, where visual occlusion (as shown in Fig. \ref{fig:force_ablation}(a)) often renders purely visual servoing unreliable. In contrast, the force-aware residual policy can interpret tactile signals to resolve misalignment and compensate for the limited precision of the visual-proprioceptive base policy. These results demonstrate that force sensing is not merely supplementary but critical for achieving the high-precision, contact-rich control necessary for robust industrial assembly.

\subsection{Discussion}
Generalizing from the experimental results, the proposed HIL-ResRL adapter demonstrates compatibility with diverse VLA models and visuomotor policies. By learning online residual corrections within a tractable local optimization scope, our method mitigates the cumulative errors and distribution shifts inherent in the base policy's inference, thereby significantly enhancing task success rates. Furthermore, this design operates without requiring access to the model's internal weights or generative paradigms (e.g., denoising or flow matching), establishing it as a strictly model-agnostic, plug-and-play solution.

%% file: conclusions.tex
\section{CONCLUSIONS AND FUTURE WORK}\label{Sec:concl}
In this work, we proposed HIL-ResRL, a model-agnostic, plug-and-play residual RL adapter for real-world deployment of vision-language-action and visuomotor models. By freezing the pre-trained base policy and learning a lightweight residual adapter with human-in-the-loop guidance, we effectively alleviate compounding errors and distribution shift from imitation learning, while ensuring safe and sample-efficient real-world training. Integrating force feedback further boosts performance in contact-rich tasks. This work provides a practical, flexible, and safe solution for deploying robotic policies in the real world. This approach represents a competitive solution for enabling rapid task switching in flexible production lines while maintaining deployable model performance. Building on this foundation, future research will extend the residual RL adapter to support multi-step long-horizon manipulation tasks in industrial scenarios.
\section*{ACKNOWLEDGMENT}